\documentclass[letterpaper]{article} 
\usepackage{aaai24}  
\usepackage{times}  
\usepackage{helvet}  
\usepackage{courier}  
\usepackage[hyphens]{url}  
\usepackage{graphicx} 
\usepackage{natbib}  
\usepackage{caption} 
\usepackage{algorithm}
\usepackage{algorithmic}

\usepackage{newfloat}
\usepackage{listings}
\DeclareCaptionStyle{ruled}{labelfont=normalfont,labelsep=colon,strut=off} 
\floatstyle{ruled}
\newfloat{listing}{tb}{lst}{}
\floatname{listing}{Listing}
\title{Precision matters: Precision-aware ensemble \\ for weakly supervised semantic segmentation}
\author{
    Junsung Park, Hyunjung Shim
}
\affiliations{
    \textsuperscript{\rm 1}Kim Jaechul Graduate School of AI\\

    Korea Advanced Institute of Science and Technology\\
    85, Hoegi-ro, Dongdaemun-gu, Seoul, Korea\\
    \{jshackist, kateshim\}@kaist.ac.kr
}

\usepackage{graphicx, subfigure, amssymb}
\usepackage{xcolor}

\begin{document}

\maketitle

\begin{abstract}
Weakly Supervised Semantic Segmentation (WSSS) employs weak supervision, such as image-level labels, to train the segmentation model. Despite the impressive achievement in recent WSSS methods, we identify that introducing weak labels with high mean Intersection of Union (mIoU) does not guarantee high segmentation performance. Existing studies have emphasized the importance of prioritizing precision and reducing noise to improve overall performance. In the same vein, we propose ORANDNet, an advanced ensemble approach tailored for WSSS. ORANDNet combines Class Activation Maps (CAMs) from two different classifiers to increase the precision of pseudo-masks (PMs). 
To further mitigate small noise in the PMs, we incorporate curriculum learning. This involves training the segmentation model initially with pairs of smaller-sized images and corresponding PMs, gradually transitioning to the original-sized pairs.
By combining the original CAMs of ResNet-50 and ViT, we significantly improve the segmentation performance over the single-best model and the naive ensemble model, respectively. We further extend our ensemble method to CAMs from AMN (ResNet-like) and MCTformer (ViT-like) models, achieving performance benefits in advanced WSSS models. It highlights the potential of our ORANDNet as a final add-on module for WSSS models.
\end{abstract}

\section{Introduction}

\begin{figure}[ht]
\begin{center}
   \subfigure{\includegraphics[width=0.9\linewidth]{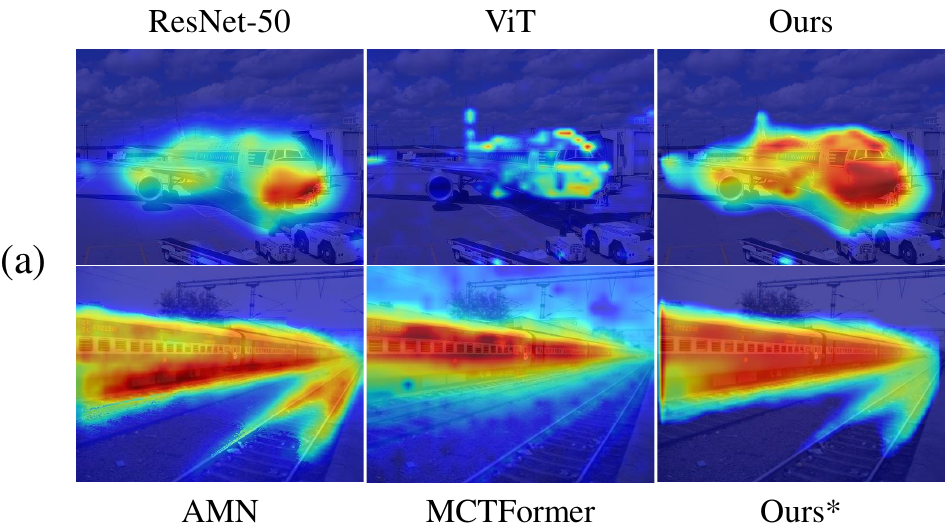}} \\
       \subfigure{\includegraphics[width=0.9\linewidth]{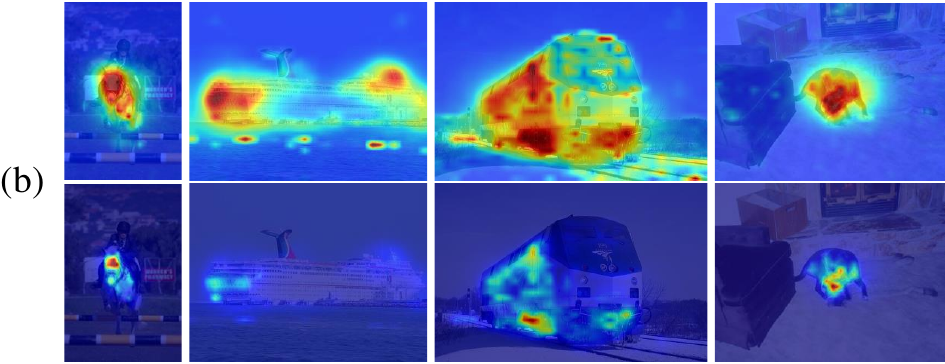}} \\
\end{center}
   \caption{(a) Difference between ResNet-based CAM and ViT-based CAM. ``Ours'' and ``Ours*" indicate the ORANDNet ensemble of ResNet-50-ViT and AMN-MCTformer, respectively. (b) Visualization of \texttt{OR} CAM (1st row) and \texttt{AND} CAM (2nd row).}
\label{fig:ResNet_vit}
\label{fig:amnmct_comparison}
\label{fig:orandcomp}
\end{figure}

Weakly supervised semantic segmentation (WSSS) addresses the high annotation cost of fully supervised semantic segmentation (FSSS) by utilizing weak supervision such as scribble \cite{lin2016scribblesup}, bounding box  \cite{khoreva2017simple}, point \cite{chen2021seminar} or image-level labels \cite{pinheiro2015image}. In this paper, we focus on WSSS methods using image-level labels since image-level labels are commonly accessible in diverse benchmarks and widely employed in practical applications. 

Existing WSSS methods using image-level labels typically follow a two-stage process. Firstly, an image classifier is trained using image-level labels, and pseudo-masks are generated through Class Activation Mapping (CAM). Subsequently, a segmentation model is trained using pairs of training images and their corresponding CAM-generated pseudo-masks. Existing methods have identified three limitations of the CAM-generated pseudo-masks: 1) the sparse coverage by capturing the most discriminative parts, 2) inaccurate boundaries due to lack of pixel-level supervision, and 3) falsely capturing the co-occurring objects.
The sparse coverage issue has been resolved by erasing the most discriminative regions of training images \cite{wei2017object, kweon2021unlocking} or consistency regularization \cite{wang2020self} or contrastive learning \cite{du2022weakly}. Predicting affinity between pixels \cite{ahn_learning_2018, ahn2019weakly} resolved the inaccurate boundary problem. Employing the saliency map \cite{lee2021railroad} and separating of object and context via data augmentation \cite{su2021context} resolved the co-occurrence problem.

Despite the impressive achievement in recent WSSS methods, we have identified that solely improving pseudo-masks in terms of mean Intersection of Union (mIoU) in the first stage does not always result in the high performance of a final model. This counter-intuitive result is due to the fact that mIoU alone may not be the best metric to evaluate the pseudo-mask quality. Recent studies on learning with noisy labels highlight the importance of precision in the early training stage via memorization effects \cite{rong2023boundary, liu2022adaptive}. Since the pseudo-mask is inevitably noisy, the importance of precision can be applied to the WSSS framework. However, the mIoU is affected by both recall and precision, and thus, the pseudo-masks with high mIoU can be achieved by increasing the recall. Since the recall does not penalize false positive predictions, the high recall inevitably allows noises in pseudo-masks, degrading the segmentation training in the second stage. This explains the discrepancy between the pseudo-masks' performance and segmentation performance, as also pointed out by \cite{rong2023boundary}.

To address the above-mentioned issue, we introduce ORANDNet, an ensemble method for WSSS that focuses on achieving high precision in pseudo-masks. Our approach leverages CAMs of two distinct classifiers to improve the precision and handle label noise within pseudo-masks.
We design ORANDNet to take a probabilistic \texttt{OR} of two CAMs as an input and to produce a probabilistic \texttt{AND} of two CAMs as an output. This simple-yet-effective approach leads to higher mIoU as well as precision in the pseudo-mask compared to using any single model or the ensemble average of two CAMs. To fully enjoy the ensemble effects of ORANDNet, we choose two distinct classifiers having clearly different characteristics, such as ResNet \cite{he2016deep} and ViT \cite{dosovitskiy2020image}. Furthermore, we confirm that incorporating more advanced WSSS methods such as AMN \cite{lee2022threshold} and MCTformer \cite{xu2022multi} can yield even higher performance, showcasing the potential of our approach to be add-on module across different WSSS models.
Additionally, We employ 'scale scheduling' early in training to downscale image-pseudo-mask pairs, removing small false activations in ORANDNet pseudo masks. This downsizing reduces noisy label impacts, thereby improving pseudo-mask precision, essential in early stages and advantageous in subsequent stages.

We believe that our work suggests the first task-specific ensemble method in WSSS. It can be an add-on module to various WSSS models, serving as a final step for future WSSS studies.

\section{Related Works}

Recent studies of WSSS often generate pseudo-labels using CAMs, derived from image-level labels, to form pixel-level pseudo-labels. Techniques like denseCRF \cite{krahenbuhl2011efficient}, AffinityNet \cite{ahn_learning_2018}, IRN \cite{ahn2019weakly}, and PAMR \cite{araslanov2020single} are used to improve boundary performance. These pseudo-labels are then fed into a Fully Supervised Semantic Segmentation Network for final training.

AdvErasing \cite{wei2017object} iteratively obscures and re-learns from a class activation map to uncover alternate discriminative regions, while CGN \cite{kweon2021unlocking} masks images with CAMs for iterative reclassification, promoting whole-object region consideration.
PPC \cite{du2022weakly} boosts WSSS performance by using contrastive learning and consistency regularization for pixel-level supervision. SEAM \cite{wang2020self} enhances CAM through an equivariance constraint, ensuring pixel class consistency after affine transformations. CDA \cite{su2021context} addresses WSSS co-occurrence by differentiating objects from their context
Fan et al. \cite{fan2020employing} propose using multiple seeds with adaptively weighted pseudo-masks per pixel. ADELE \cite{liu2022adaptive} corrects labels by optimally weighting them during early, less noise-prone learning epochs. BECO \cite{rong2023boundary} enhances noise resistance in segmentation networks using a Siamese network for uncertain regions and creating mixed image-label-boundary pairs.
WeakTr \cite{zhu2023weaktr} introduces an adaptive attention fusion module for weighting across attention heads, deriving class activation maps. RCA \cite{zhou2022regional} uses contrastive learning to understand regional priors across datasets, surpassing individual image prior dependency.


\section{Methods}

\begin{figure}[t]
\begin{center}
   \subfigure{\includegraphics[width=\linewidth]{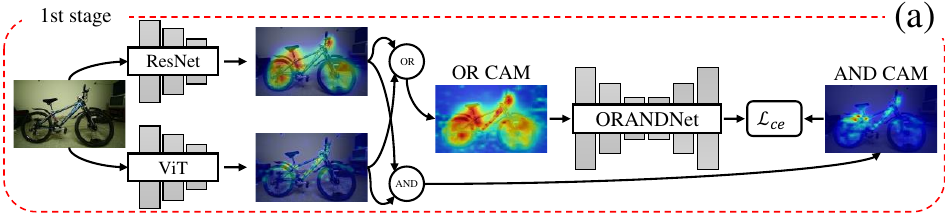}} \\
   \subfigure{\includegraphics[width=\linewidth]{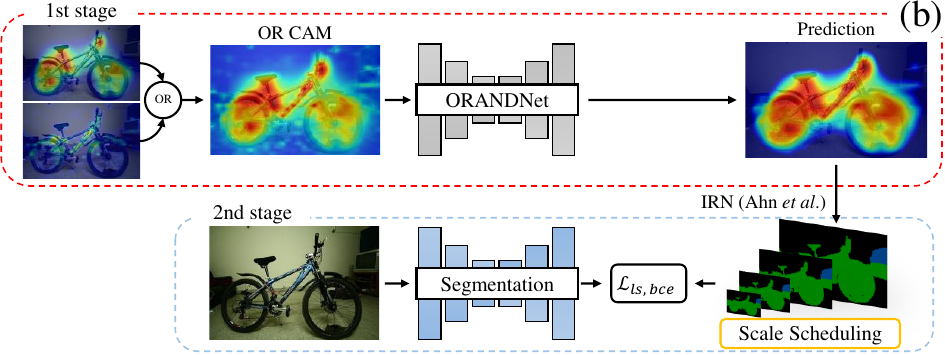}} \\
\end{center}
   \caption{The overall pipeline of our method includes: (a) ORANDNet during training, and (b) ORANDNet in the test time with scale scheduling. $\mathcal{L}_{\textit{ce}}$ and $\mathcal{L}_{\textit{ls, bce}}$ mean cross entropy and class-balanced cross entropy with label smoothing, respectively.}
\label{fig:orandtrain_small}
\label{fig:fullprocess}
\end{figure}

This section provides a detailed procedure for ORANDNet. Unlike existing methods, ORANDNet takes two CAMs from two different classifiers and performs an effective ensemble aimed at boosting precision. The comprehensive process is exhibited in Figure ~\ref{fig:fullprocess}. In the following, we first discuss the rationale behind ORANDNet, outline the overall process, and demonstrate ORANDNet with its training and architecture.

\subsection{Motivation}

To fully exploit the benefits of ensemble methods, it is important to use two distinct Class Activation Maps (CAMs), where each possesses distinguishable characteristics compared to another. With this objective in mind, we employ ResNet and Vision Transformer (ViT) for their unique CAM characteristics: ResNet CAMs show concentrated activations at key object points, while ViT's are denser but weaker, covering more of the object (see Figure~\ref{fig:ResNet_vit} (a)). Both exhibit false activations beyond object boundaries.
Activation map inconsistencies presented between ResNet and ViT extend to WSSS models with these backbones. Figure \ref{fig:amnmct_comparison} (a) illustrates this, contrasting AMN (based on ResNet-38) and MCTformer (a ViT variant).

Our goal is to develop a task-specific ensemble method that enhances pseudo-mask precision. Leveraging the constrasting distributions in activation maps of ResNet and ViT, we identify consistent activations as reliable pseudo-labels. Thus, a probabilistic \texttt{AND} operation on these CAMs results in a map with reduced false activations, exemplified in Figure ~\ref{fig:orandcomp} (b) using the ResNet-ViT pair.

While consistent activations contribute to improved precision in pseudo masks, achieving sufficient mIoU scores also requires reasonable recall. Training a segmentation model with high precision but low recall is problematic due to data scarcity. Consequently, the result of the AND operation, as shown in Figure ~\ref{fig:orandcomp} (b), is not conducive to effective segmentation training. To achieve high mIoU and precision simultaneously, we propose ORANDNet.

\subsection{Proposed method}

Following the convention of existing WSSS methods, the pipeline of the proposed method consists of two stages. The first stage involves obtaining the pseudo-mask. The second stage uses pairs of images and pseudo-masks to train the semantic segmentation network. The detailed procedures are illustrated in Figure ~\ref{fig:fullprocess}.

\subsubsection{ORANDNet}

As mentioned in the previous section, our goal is to enhance pseudo-mask precision and reduce noise, without compromising on the quality of the pseudo-masks.
First, we execute pixel-wise probabilistic \texttt{OR} and \texttt{AND} operations on the two CAMs from distinct models. \texttt{OR} and \texttt{AND} operations are illustrated as below respectively:

\begin{equation}
    {\small \displaystyle\frac{c_1 + c_2 - c_1 \circledcirc c_2}{\max(c_1 + c_2 - c_1 \circledcirc c_2)}}, \space
    {\small \displaystyle\frac{c_1 \circledcirc c_2}{\max(c_1 \circledcirc c_2)}}.
\label{eqn:cam_or_and}
\end{equation}
where $\circledcirc$ is a pixel-wise multiplication, and $c_1$ and $c_2$ indicate two distinctive CAMs from different models. By applying Equation ~\ref{eqn:cam_or_and}, we can derive the \texttt{OR} CAM and \texttt{AND} CAM.

As previously discussed, training a segmentation model using pseudo-masks generated from the \texttt{AND} CAM is hindered by data scarcity, cause by low recall of pseudo-mask.
To address this challenge and produce high-precision pseudo-masks, we propose a simple yet effective approach called ORANDNet. In this method, we design an FCN \cite{simonyan2014very} to predict the \texttt{AND} CAM from the \texttt{OR} CAM, which is derived from two distinct activation maps.
ORANDNet improves pseudo-mask precision and mIoU by refining dense, wide-coverage activation maps into consistent regions while retaining sufficient recall, with Figure ~\ref{fig:ResNet_vit} (a) demonstrating these enhancements.

After achieving \texttt{OR} CAM and \texttt{AND} CAM through Eqn ~\ref{eqn:cam_or_and}, ORANDNet is then trained with the \texttt{AND} CAM as the pseudo-mask and the \texttt{OR} CAM as the input. In this way, ORANDNet learns to reduce noise in the pseudo-masks and enhance precision while preserving the object coverage of activations.
During the testing phase, the \texttt{OR} CAM is utilized as the input for ORANDNet. Subsequently, we achieve the pseudo-mask by applying the IRN to the predictions of ORANDNet.

\subsubsection{Scale scheduling}
To reduce noise in pseudo-masks, we adopt a curriculum learning strategy in second stage, progressively downsampling training data (image and pseudo-mask pairs) by factors of 8, 4, and 2 for the initial 2, 4, and 6 epochs, respectively. This method effectively discards smaller noise early in training by introducing variably downsampled image-pseudo mask pairs at different epochs.

\section{Experiments}

\noindent\textbf{Datasets and evaluation metrics.} The PASCAL VOC 2012 dataset~\cite{everingham2015pascal} was used for these experiments. The training set of PASCAL VOC 2012 comprises 10,582 images, the validation set contains 1,449 images, and the test set contains 1,456 images.
We evaluated pseudo-mask generation using mean Intersection over Union (mIoU) and precision metrics, while segmentation results were assessed solely based on mIoU.

\noindent\textbf{Implementation details.} When training ResNet-50 with classification, we used a learning rate of 0.01, a polynomial scheduler, and a weight decay of 5e-4. For ViT, the learning rate was set to 0.001, with a cosine annealing scheduler and a weight decay of 1e-5. ORANDNet used a learning rate of 0.001, a polynomial scheduler, and no weight decay. ORANDNet utilizes the VGG16 backbone without batch normalization \cite{simonyan2014very}. 
For the final segmentation model, the Deeplab-v1 model \cite{chen2014semantic} with a ResNet-38 backbone \cite{wu2019wider} was selected.

\subsection{Ablation study}

\begin{table}
\begin{center}
\scalebox{0.8}{
\begin{tabular}{|l|c|c|c|}
\hline
Method & mIoU & Precision & Recall \\
\hline\hline
ResNet-50 & 48.3 & 66.9 & 64.8 \\
ViT & 53.4 & 67.7 & 72.4 \\
Na\"ive ensemble & 49.0 & 61.5 & 72.2 \\
\textbf{Ours} & \textbf{54.3} \textcolor{red}{\small(+5.0)} & \textbf{71.0} \textcolor{red}{\small(+9.5)} & 70.9 \textcolor{red}{\small(-1.3)}\\
\hline
\textbf{Ours w/IRN} &\textbf{70.9} & \textbf{82.9} & 82.6 \\
\hline \hline
AMN & 62.8 & 74.0 & 80.5 \\
MCTformer & 62.1 & 74.7 & 78.8 \\
\textbf{Ours*} & \textbf{64.3} \textcolor{red}{\small(+1.5, +2.2)} & \textbf{78.9} \textcolor{red}{\small(+4.9, +4.2)} & 77.9 \textcolor{red}{\small(-2.6, -0.9)} \\
\hline
\textbf{Ours* w/IRN} & \textbf{74.3} & \textbf{85.5} & 84.3 \\
\hline
\end{tabular}
}
\end{center}
    \caption{The results of pseudo-masks for ORANDNet and na\"ive ensembles, and single models. Classifier training was done on the PASCAL VOC 2012 \textit{augmented train set}, evaluated on the \textit{train set}. Increments were based on the na\"ive ensemble for `Ours', AMN/MCTFormer for `Ours*'.}
\label{tab:resvit_simpleavg}
\label{tab:miou_resvit}
\label{tab:amnmct}
\end{table}

\noindent\textbf{ORANDNet with basic classification models.} Firstly, we investigate the effect of ORANDNet using two classifiers, ResNet-50 and ViT-base-patch16.

Table ~\ref{tab:miou_resvit} summarizes the mIoU, precision, and recall of ORANDNet compared to those of CAM with ResNet and ViT, respectively. The results clearly show that ORANDNet improves both mIoU and precision better than the individual model. 
Specifically, by applying ORANDNet, we gain +6 in mIoU and +4.1 in precision relative to ResNet-50. Similarly, with the ViT backbone, we achieve +0.9 in mIoU and +3.3 in precision. These results support that ORANDNet is clearly effective in enhancing precision and mitigating noise. Finally, employing IRN further improves both mIoU and precision.

\noindent\textbf{Add-on on other models.} Beyond ResNet and ViT, we explored ORANDNet's performance enhancement with other high-performing WSSS models, specifically AMN (a ResNet-50 variant) and MCTFormer (a ViT variant). We generated pseudo-masks using both AMN and MCTFormer, integrating them with ORANDNet, akin to our earlier experiments.
As shown in row 8 of Tabel ~\ref{tab:amnmct}, ORANDNet gain +1.5 mIoU/+4.9 precision and +2.2 mIoU/+4.2 precision respect to AMN and MCTFormer.
These results emphasize the effectiveness of the CAM/pseudo-mask ensemble approach of ORANDNet on advanced WSSS methods. It shows its potential as a final add-on module for future WSSS models.

\noindent\textbf{Comparison with a na\"ive ensemble.} Effectiveness of ORANDNet was compared against a naïve ensemble, which simply averages CAMs from ResNet and ViT. Results in row 4 of Table ~\ref{tab:resvit_simpleavg} show +5.0 mIoU and +9.5 precision gain in respect to na\"ive ensemble. These lead to two conclusions: 1) Ensemble approaches improve pseudo mask quality, and 2) ORANDNet outperforms the naïve ensemble.


\subsection{Comparing segmentation results with recent WSSS}

In these experiments, we utilized pseudo-masks for training a segmentation model and evaluated the performance of recent WSSS methods on the PASCAL VOC 2012 \textit{validation set} and \textit{test set}.

\noindent\textbf{Quantitative results.} As in Table ~\ref{tab:recentwsss}, we found that precision improved pseudo-mask from ORANDNet can yield  comparable performance to state-of-the-art WSSS methods with ResNet-50 and ViT.
This indicates that enhancing the precision of pseudo-masks contributes significantly to the enhancement of segmentation quality

In contrast to the methodological complexity of existing WSSS methods and their limited gains, our simpler method with ResNet-50 and ViT gained +6.8/+7.3 mIoU respect to IRN(with ResNet-50) on validation/test set. Considering that +1.0 mIoU gain was often seen in prior WSSS studies, +6.8/+7.3 mIoU gain using the simple method we propose is a significant achievement.
To highlight this achievement, our method with AMN and MCTFormer achieved an increase of +2.3 and +1.3 in mIoU on the test set, respectively, compared to the performance of each model when used individually.

Note that the second stage methods (denoted as `2nd' in Table ~\ref{tab:recentwsss}) suggest the segmentation training strategy thus they can be easily combined with the first stage methods (denoted as `1st'). That means, our method can be used in conjunction with BECO and further improve their performance.
Additionally, for future WSSS methods that produce more accurate pseudo-masks, incorporating ORANDNet into those approach has the potential to further enhance its performance.

\begin{table}[t]
\begin{center}
\scalebox{0.75}{
\begin{tabular}{|l|c|c|c|c|}
\hline
Method & Stg & Sup. & Val & Test \\
\hline\hline
EDAM \cite{wu2021embedded} & 1st & I+S & 70.9 & 70.6 \\
EPS \cite{lee2021railroad} & 1st & I+S & 71.0 & 71.8 \\
SANCE \cite{li2022towards} & 1st & I+S & 72.0 & 72.9 \\
\hline
OC-CSE \cite{kweon2021unlocking} & 1st & I & 68.4 & 68.2 \\
CDA \cite{su2021context} & 1st & I & 66.1 & 66.8 \\
PPC \cite{du2022weakly} & 1st & I & 72.6 & \textbf{73.6} \\
\hline
URN \cite{li2022uncertainty} & 2nd & I & 69.5 & 69.7 \\
ADELE \cite{liu2022adaptive} & 2nd & I & 69.3 & 68.9 \\
BECO \cite{rong2023boundary} & 2nd & I & \textbf{73.7} & 73.5 \\
\hline \hline
IRN \cite{ahn2019weakly} & 1st & I & 63.5 & 64.8 \\
Ours & 1st & I & 70.3 & 72.1 \\
\hline
\small{Relative to IRN} &  &  & \textcolor{red}{\small{+6.8}} & \textcolor{red}{\small{+7.3}}\\
\hline\hline
AMN \cite{lee2022threshold} & 1st & I & 70.7 & 70.6 \\
MCTformer \cite{xu2022multi} & 1st & I & 71.9 & 71.6 \\
Ours* & 1st & I & 72.2  & 72.9 \\
\hline
\small{Relative to AMN, MCTFormer} &  &  & \textcolor{red}{\small{+1.5, +0.3}} & \textcolor{red}{\small{+2.3, +1.3}}\\
\hline
\end{tabular}
}
\end{center}
    \caption{mIoU of segmentation results on PASCAL VOC 2012 validation/test set. Ours/Ours* are ORANDNet ensemble of ResNet-50/ViT and AMN/MCTformer, respectively. Sup. denotes the weak supervision type, I is image-level labels and S is saliency supervision. Stg. denotes stages developed in each method.}
\label{tab:recentwsss}
\end{table}

\begin{figure}
\begin{center}
    \includegraphics[width=\linewidth]{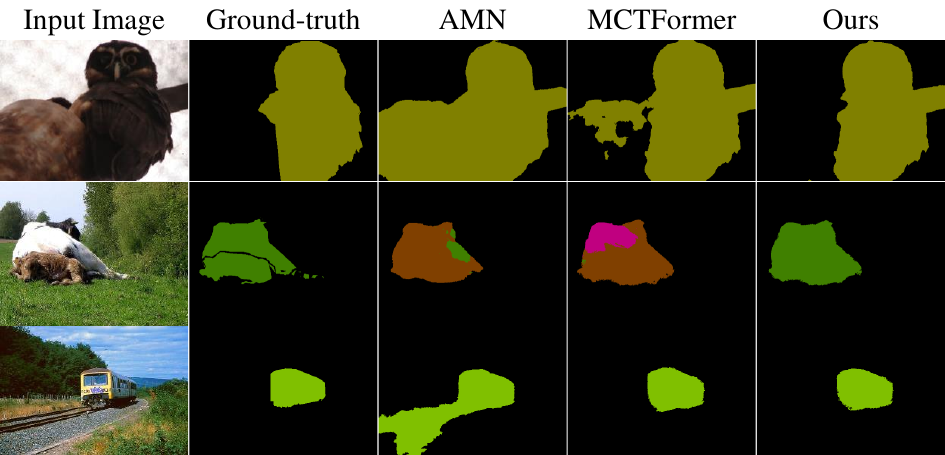}
\end{center}
   \caption{Segmentation results of AMN and MCTformer and their ORANDNet ensemble.}
\label{fig:final_result_amnmct}
\end{figure}

\noindent\textbf{Qualitative results.} Figure ~\ref{fig:final_result_amnmct} visualizes the improved segmentation results from ORANDNet's pseudo-masks. The first row highlights improved precision by our method. The second row highlights ORANDNet's capability to correct false predictions within single objects, which is not observed with existing methods. In the third row, we observe that ORANDNet retains the advantage of existing methods in preventing co-occurrence, inherited from MCTformer. Collectively, these three results demonstrate ORANDNet's simplicity-yet-powerful ability in enhancing WSSS performance.

\section{Conclusion}
We present ORANDNet, a novel ensemble approach in WSSS focusing on pseudo-mask precision. It markedly improves mIoU and precision, acheiving performance on par with state-of-the-art even with basic classifiers like ResNet-50 and ViT. As the first WSSS-specific ensemble method, ORANDNet can be employed as add-on to future WSSS methodologies.

\section{Achknowledgement}
This research was supported by 
the Basic Science Research Program through the National Research Foundation of Korea (NRF) funded by the MSIP (NRF-2022R1A2C3011154, RS-2023-00219019)
and IITP grant funded by the Korea government(MSIT) and KEIT grant funded by the Korea government(MOTIE) (No. 2022-0-01045, No. 2022-0-00680).


\bibliography{aaai24}

\end{document}